\documentclass[submission]{eptcs}
\providecommand{\event}{ICLP 2020} 
\usepackage{breakurl}             
\usepackage{underscore}           
\usepackage{amsmath}
\usepackage{graphicx}
\usepackage{hyperref}
\usepackage{mathptmx}
\usepackage{pgfplots}
\usepackage{subcaption}
\usepackage{xcolor}
\usepackage{xspace}
\pgfplotsset{width=8.2cm,height=5cm,compat=1.9}

\newcommand\tabdual{\textsc{Tabdual}\xspace}

\newtheorem{theorem}{Theorem}[section]
\newtheorem{definition}{Definition}[section]
\newtheorem{example}{Example}[section]
\newtheorem{experiment}{Experiment}[section]
\DeclareMathAlphabet{\mathcal}{OMS}{cmsy}{m}{n}

\title{Tabling Optimization for Contextual Abduction}
\author{Ridhwan Dewoprabowo \qquad\qquad Ari Saptawijaya
\institute{Faculty of Computer Science\\
Universitas Indonesia\\
Depok, Indonesia}
\email{\quad ridhwan.dewoprabowo@ui.ac.id \quad\qquad saptawijaya@cs.ui.ac.id}
}
\def\titlerunning{Tabling Optimization for Contextual Abduction}
\def\authorrunning{R. Dewoprabowo \& A. Saptawijaya}

\begin{document}
\maketitle

\begin{abstract}
Tabling for contextual abduction in logic programming has been introduced as a means to store previously obtained abductive solutions in one context to be reused in another context. This paper identifies a number of issues in the existing implementations of tabling in contextual abduction and aims to mitigate the issues. We propose a new program transformation for integrity constraints to deal with their proper application for filtering solutions while also reducing the table memory usage. We further optimize the table memory usage by selectively picking predicates to table and by pragmatically simplifying the representation of the problem. The evaluation of our proposed approach, on both artificial and real world problems, shows that they improve the scalability of tabled abduction compared to previous implementations.
\end{abstract}

\section{Introduction}
\label{sect:introduction}

The requirement for artificial intelligence (AI) to provide explanations in making critical decision becomes increasingly important due to concerns of accountability, trust, as well as ethics. Such an explainable AI is expected to be capable of providing justifications that are human-understandable. A form of reasoning for providing explanations to an observation, known as \textit{abduction}, has been well studied in AI, particularly in knowledge representation and reasoning. It extends to logic programming, dubbed \textit{abductive logic programming} \cite{DK02}, and it has a wide variety of usage, e.g., in planning, scheduling, reasoning of rational agents, security protocols verification, biological systems, and machine ethics. \cite{AC05,GT00,KM01,KS11,LK12,PS16}.

Finding best explanations in abduction can be very costly, especially when the knowledge base is large and complex. In some cases, abductive solutions already obtained within one abductive context may be relevant to other contexts and thus, can be reused in those contexts. This idea of storing and reusing abductive solutions across different contexts, designated as \textit{contextual abduction}, has been recently brought into abductive logic programming \cite{SP15}. It benefits from tabling mechanism in logic programming \cite{swift99-tabling}, which is supported to different extents in a number of Prolog systems. Tabling for contextual abduction, \tabdual, is implemented on top of XSB Prolog \cite{sw12-xsb}, and is realized by a transformation from abductive logic programs into tabled logic programs. The transformation makes use of the dual program transformation \cite{AP04} to deal with abduction under negative goals.

While several implementation aspects have been considered in \cite{SP15}, \tabdual may still suffer from excessive computational cost, particularly in terms of space due to tabling itself. This is especially true for an observation where many and large alternative explanations are found and naively tabled. Consequently, it may hamper contextual abduction to be completed, as it requires too many resources before being able to return a solution. To address this scalability issue, various features of tabling mechanism have been employed. In \cite{PS17}, \tabdual is extended with answer subsumption \cite{SW10} to store only subsuming abductive solutions and delivers to that effect only subset-minimal explanations to an observation. Using a real world case of chemoprevention \cite{LK12} as a benchmark, answer subsumption improves \tabdual in computing (minimal) abductive solutions for a query with more goals, as the number of solutions is reduced. Nevertheless, scalability remains a challenge for harder queries (i.e., queries built from many goals) of the same benchmark. A way around this challenge is to call queries incrementally with respect to the number of their goals, making the most of tabled solutions from simpler queries (with less goals) and reusing them to answer harder ones.

In another exploration \cite{IS19}, a more recent tabling feature in XSB Prolog, viz., tabling with interned ground terms \cite{DW13}, is exploited to improve the scalability of \tabdual. As abductive solutions are assumed to be ground, they can benefit from the succinct representation of interned ground terms and, at the same time, they are tabled for future reuse in other contexts. Unfortunately, in XSB Prolog, tabling with interned ground terms makes use of variant tabling, and thus cannot be combined with answer subsumption. It can therefore complement the variant of \tabdual that employs answer subsumption, viz., to provide all solutions rather than only minimal ones. Indeed, experimenting with the same benchmark demonstrates that, when it is compared with the one without interning operation, it is superior in the case of returning all solutions, though it takes longer time to return the first solution. Nevertheless, this variant of \tabdual still struggles in solving the hard queries of the same benchmark.

With the motivation of improving the scalability of \tabdual, furthering the existing implementations that have exploited various tabling features, in this study we focus on the program transformation of \tabdual. We look into the abductive solutions produced by the transformation and determine whether or not these solutions should be tabled. In so doing, we identify that integrity constraints should be handled carefully in order for \tabdual to correctly get rid of unwanted solutions but without rejecting valid solutions. The structure of rules is also taken into account in deciding which predicates to table.

The contributions of the paper are therefore as follows. First, we formulate a new transformation for integrity constraints to warrant their correct application in filtering solutions. This new transformation no longer depends on the dual program transformation, but instead it utilizes subset checking as a way to ensure that abductive solutions adheres to the integrity constraints. Second, we introduce a mechanism to reduce the number of tabled solutions. This is realized by selectively choosing predicates to table, e.g., by inspecting the occurrence of abducibles in the body of rules. We design an artificial problem, whose size is parameterized by the number and the size of rules, to study the table memory usage of the proposed approach. The results show that, compared to the implementation where the present approach is not in place, the table memory usage is reduced as the size of the problem increases. However, as an implication, the number of inferences may also increase. There is therefore a trade-off between the table memory usage and the number of inferences, as expected. In the end, the scalability of \tabdual with the present approach is evaluated on the same real world case as in the previous implementations. Together with answer subsumption, the new transformation for integrity constraints and the selection of tabled predicates successfully reduce the table memory usage compared to the implementation that only employs answer subsumption. Consequently, compared to the previous implementation with answer subsumption \cite{PS17}, the present approach manages to solve queries with more goals,  even when queries are not called incrementally. We also exploit the representation of the problem and simplify the representation to further reduce the table memory usage.

The rest of the paper is organized as follows. We set up notations and recap contextual abduction in logic programming, in Section \ref{sect:prelim}. The new transformation for integrity constraints and its empirical analysis are explained in Section \ref{sect:ic}. Section \ref{sect:scale} discusses our proposed method for optimizing table memory usage by reducing tabled predicates and its evaluation on artificial and real world cases. We  conclude, in Section \ref{sect:cfw}, with future work. 
The implementation code for our improvement of \tabdual is available on \url{https://github.com/RidhwanD/TabdualSC}.

\section{Contextual Abduction in Logic Programming}
\label{sect:prelim}
	
A \textit{logic program} $\mathcal{P}$ is a countable set of rules in the form $H \leftarrow L_1, \dots, L_m$, $m \ge 0$ where $L_i$, $1 \le i \le m$ are literals (atoms or their negation). Ground terms (literals, rules, and programs) are defined as usual, without variables.

Abduction is a form of reasoning that aims at finding plausible explanations to a given observation. In logic programming, abduction is implemented by extending the logic program with abductive hypotheses or \textit{abducibles} as candidates of explanation to an observation; the latter is  given as a query. We recap below abductive reasoning in well-founded semantics \cite{VR91} and closely follow the definitions in \cite{AP04}. We use $\bot$ and $\textsf{U}$ to denote \textit{false} and \textit{undefined}, respectively. In abductive reasoning, \textit{integrity constraints} (ICs) are rules in the form of denial $\bot \leftarrow L_1,\dots,L_m$. We define $\mathcal{AB}$ as the set of abducible predicates, i.e., predicates without rules. We define $\mathcal{AB}_g$ as the set containing ground abducible literals, i.e., for each $ab \in \mathcal{AB}$, we have that $ab(\bar{t})$, $\textit{not } ab(\bar{t}) \in \mathcal{AB}_g$ for all $\bar{t}$ tuples of ground terms. We say that $\textit{not } ab(\bar{t})$ is the complement of $ab(\bar{t})$ and vice versa.

\begin{definition} [Abductive Framework]
An abductive framework $F$ is a triple $\langle \mathcal{P}, \mathcal{AB}, \mathcal{IC} \rangle$ where $\mathcal{P}$ is a logic program such that no rule in $\mathcal{P}$ has an abducible from $\mathcal{AB}$ as its head, and $\mathcal{IC}$ is a set of integrity constraints.
\end{definition}

\begin{definition} [Abductive Solution] \label{abdsol}
	Given an abductive framework $F = \langle \mathcal{P}, \mathcal{AB}, \mathcal{IC} \rangle$, the set of abducibles $\mathcal{S} \subset \mathcal{AB}_g$ is an abductive solution for $F$ if $\bot$ is false in well-founded model $M_{\mathcal{S}}$ of $(\mathcal{P} \cup \mathcal{P}_{\mathcal{S}} \cup \mathcal{IC})$ where $\mathcal{P}_{\mathcal{S}}$ is the smallest set of rules containing the fact $A$ if $A \in \mathcal{S}$ and, for all $A' \in \mathcal{AB}_g \backslash \mathcal{S}$ and $A'$ is not the complement of $A$, we have that $A' \leftarrow \textsf{U}$. A set $\mathcal{S}$ is an abductive solution for query $Q$ if $M_{\mathcal{S}} \models Q$.
\end{definition}

\begin{example} \label{ex1}
    Given the abductive framework $\langle \mathcal{P}, \mathcal{AB}, \mathcal{IC} \rangle$ where $\mathcal{AB}=\{q, r, t\}$ and $\mathcal{P}$ is the program:
    \begin{align*}
        p(X) &\leftarrow q(0), q(1), s(X).  & \hspace{0.5cm}
        s(X) &\leftarrow \textit{not } t(X). & \hspace{0.5cm}
        u(X) &\leftarrow \textit{not } p(X).
    \end{align*}
    and $\mathcal{IC} = \{\bot \leftarrow q(X), r(X), \bot \leftarrow u(X)\}$. Given a query $Q = \{ p(0) \}$, its abductive solution is $\mathcal{S} = \{q(0), q(1), \textit{not } t(0), \textit{not } r(0), \textit{not } r(1)\}$. \\ Note that $\{q(0), q(1), \textit{not } t(0), \textit{not } r(0), \textit{not } r(1), \allowbreak s(0), p(0), \textit{not } u(0)\} \subset M_{\mathcal{S}}$ and thus $M_{\mathcal{S}} \models Q$.
\end{example}

\textit{Contextual abduction} in \tabdual \cite{SP15} implements tabling mechanism for abduction in well-founded semantics to table abductive solutions obtained within one abductive context to be reused in other relevant contexts. Contexts can be the results obtained from previous queries or subgoals. Calling $s(0)$ before the query $Q$ in Example \ref{ex1} allows us to reuse its solution, i.e., $\textit{not }t(0)$, in the execution of $Q$. Context can also be seen as a way to restrict the solutions by determining the initial hypothesis of a query. In Example \ref{ex1}, calling the query $p(0)$ with context $r(0)$, i.e., $r(0)$ is assumed true when we call the query, may reuse the already computed solution obtained from query $p(0)$ without context, i.e., $\{q(0), q(1), \textit{not } t(0), \textit{not } r(0), \textit{not } r(1)\}$ by adding the context $r(0)$ into it. However, since $\textit{not } r(0)$ is present in the solution due to the first IC, the query $p(0)$ with context $r(0)$ has no solution. Tabled abduction in \tabdual involves a program transformation, that transforms abductive logic program to tabled logic program, and the abduction stage on top of the tabled logic program. The transformation introduces two extra arguments, for all predicates in the program, to serve as the input and output contexts. The program transformation process of \tabdual includes: 
\begin{enumerate}
    \item Transformation for tabling abductive solution that accommodates storing and reusing abductive solutions by utilizing XSB Prolog’s tabling mechanism. For every rule $r$, we define the rule $r_{ab}$ as its tabled predicate and moving the abducibles in its body as its input context. For each predicate $p$, we define a rule that permits reusing tabled abductive solutions in $p_{ab}$ consistently in another context.
    \begin{example}
        Given program $\mathcal{P}$ in Example \ref{ex1}. The results of transforming $p/1$ are as follow:
        \begin{itemize}
            \item $p_{ab}(X,O) \leftarrow s(X,[q(0),q(1)],O).$
            \item $p(X,I,O) \leftarrow p_{ab}(X,E), produce\_context(O,I,E).$
        \end{itemize}
        The transformation for $s/1$ and $u/1$ are defined similarly.
    \end{example}
    Here, $produce\_context(O,I,E)$ is a \tabdual system predicate that consistently produces output $O$ from the input context $I$ and a tabled solution $E$. It checks whether each abducible in $E$ or its negation is already present in $I$. If that is not the case, then the abducible is added into the context. $O$ is the resulting output context from this predicate. Otherwise, if the negation is present, then $E$ and $I$ are inconsistent and the predicate fails. 
    \item Transformation for generating dualized negation that enables \tabdual to deal with abduction under negative goals. It employs the dual program transformation from \textsc{Abdual} \cite{AP04} to deal with negative goals as ‘positive’ literals. For every predicate $p(\bar{X})$, we define the rule $not\_p(\bar{X},T_0,T_1)$ with $p^{*i}(\bar{X},T_{i-1},T_i), 1 \le i \le n$ as its body, where $n$ is the number of rules with head $p$. Each $p^{*i}(\bar{X},T_{i-1},T_i)$ represents a rule that falsifies the original $i$-th rule of $p$.
    \begin{example} \label{ex:ic}
        Given program $\mathcal{P}$ in Example \ref{ex1}, the results of transformation for $p/1$ and the ICs are defined below. The transformation for $s/1$ and $u/1$ are defined similarly.
        \begin{itemize}
            \item $p^{*1}(X,I,O) \leftarrow not\_q(0,I,O).$
            \item $p^{*1}(X,I,O) \leftarrow q(0,I,T), not\_q(1,T,O).$
            \item $p^{*1}(X,I,O) \leftarrow q(0,I,T), q(1,T,S), not\_s(X,S,O).$
            \item $not\_p(X,I,O) \leftarrow p^{*1}(X,I,O).$
            \item $false^{*1}(I,O) \leftarrow not\_q(X,I,O).$
            \item $false^{*1}(I,O) \leftarrow q(X,I,T), not\_r(X,T,O).$
            \item $false^{*2}(I,O) \leftarrow not\_u(X,I,O).$
            \item $not\_false(I,O) \leftarrow false^{*1}(I,O),false^{*2}(I,O).$
        \end{itemize}
    \end{example}
    \item Transformation for inserting abducibles into context while also maintaining the consistency of the abductive context when an abducible is abduced. Each abducible $a(\bar{X})$ transforms into two rules, $a(\bar{X},I,O)$ and $not\_a(\bar{X},I,O)$. 
    \begin{example} \label{ex:abd}
        Given the set $\mathcal{AB}$ from Example \ref{ex1}. The transformation result for $q$ are as follow:
        \begin{itemize}
            \item $q(X,I,O) \leftarrow insert\_abducible(q(X),I,O).$ 
            \item $not\_q(X,I,O) \leftarrow insert\_abducible(\textit{not } q(X),I,O).$
        \end{itemize}
        The \tabdual system predicate $insert\_abducible(A,I,O)$ non-redundantly adds abducible $A$ into context $I$ to obtain consistent context $O$. The transformation for abducibles $r$ and $t$ are defined similarly.
    \end{example}
    \item Query transformation that ensures any abductive solution satisfies the ICs. Each query is transformed by adding input and output contexts to each subgoal. To warrant that ICs are not violated,  $not\_false(I,O)$ is added at the end. The output context $O$ therefore serves as an abductive solution to the query, by checking whether $I$ satisfies the ICs.
    \begin{example} \label{ex:query}
        The query $Q$ in Example \ref{ex1} transforms into $?- p(0,[\;],T), not\_false(T,O)$ and returns $O=[q(0), q(1), \textit{not } t(0), \textit{not } r(0)]$ as a solution. Calling the query $Q$ with an input context $r(0)$ amounts to calling  $?- p(0,[r(0)],T), not\_false(T,O)$. Invoking subgoal $p(0,[r(0)],T)$ results in $T=[r(0), q(0), q(1), \textit{not } t(0)]$. Invoking the second subgoal $not\_false([r(0), q(0), q(1), \allowbreak \textit{not } t(0)], O)$ fails as $T$ violates the first IC.
    \end{example}
\end{enumerate}

\section{Transforming Integrity Constraints}
\label{sect:ic}

In the original \tabdual, ICs are maintained by first transforming them into their dual rule and conjoining it with the given query, as illustrated above. We highlight two problems concerning the dual program transformation for ICs. Firstly, the resulting $not\_false(I,O)$ adds more abducibles to $I$ for the abductive solution $O$ to satisfy ICs under well-founded semantics (according to Definition \ref{abdsol}). As a result, it may lead to a large number of generated abductive solutions, since an IC with multiple literals in the body produces several rules after transformation. Secondly, it may cause the abduction process to return incorrect solutions. For example, using the abductive framework in Example \ref{ex1}, the query $p(0)$ with the input context $r(1)$ returns the solution $[q(0), q(1), \textit{not } t(0), \textit{not }r(0), r(1)]$, despite violating the ICs. It happens since the call of $false^{*1}/2$ in line 6 of Example \ref{ex:ic} attempts to abduce $q(X)$ and $\textit{not } r(X)$ using the rules in Example \ref{ex:abd}. However, when attempting to abduce $q(X)$, $insert\_abducible/3$ already unifies $q(X)$ with the atom $q(0)$ in the solution, and additionally $not\ r(0)$ is abduced. Consequently, $q(X)$ never unifies with $q(1)$ and the consistency checking for $X=1$ is never performed. Thus, it returns the incorrect solution, where $q(1)$ and $r(1)$ belong to this solution.

The dual transformation of ICs may also risk rejection of correct solutions. Suppose that in Example \ref{ex1}, the rule $p/1$ is slightly modified into $p(X) \leftarrow q(0), \textit{not } q(1), s(X).$ Given the query $Q=\{p(0)\}$, abduction fails even though $S = [q(0), \textit{not } q(1), \textit{not } t(0)]$ is actually a solution that does not violate the IC. In this case, it fails since calling $false^{*1}/2$ in line 5 of Example \ref{ex:ic} attempts to abduce $\textit{not } q(X)$. The $insert\_abducible/3$ predicate in the rule on line 2 of Example \ref{ex:abd}, for the sake of consistency, checks whether the complement of $\textit{not } q(X)$, viz., $q(X)$, exists in $S$. Unfortunately, the variable $X$ in $q(X)$ is already unified to $0$, and consequently it fails. Alternatively, calling $false^{*1}/2$ in line 6 of Example \ref{ex:ic} also fails as it attempts to abduce $q(X)$ for similar reason. In this case, $X$ is already unified with $1$ when it checks its complement $\textit{not } q(X)$.

A way to resolve these problems is by modifying the transformation for ICs such that it avoids adding more abducibles into a solution when checking for consistency. The idea is to initially compute the lists of abducibles that satisfy the ICs and to save each list using a new predicate $ic/1$. Then, in the abduction stage, every generated abductive solution is tested against those lists by subset checking. If none of these lists is a subset of the solution, then the abductive solution does not violate the ICs. To realize this idea, the definition of IC is modified into a rule of the form $\textsf{U} \leftarrow L_1,\dots,L_m$, i.e., abducing abducibles in the body is not forced and therefore they can be left undefined. In other words, this definition allows the body of an IC to be false or undefined \cite{PA91}, allowing us to define abductive solutions under the well-founded semantics as defined below.\footnote{The use of $\textsf{U}$ in ICs is also adapted for abduction in the weak completion semantics \cite{SH18}.} 

\begin{definition} [Abductive Solution (Modified)] \label{abdsolmod}
	Given an abductive framework $F = \langle \mathcal{P}, \mathcal{AB}, \mathcal{IC} \rangle$, the set of abducibles $\mathcal{S} \subset \mathcal{AB}_g$ is an abductive solution for $F$ if $\textsf{U}$ is undefined in well-founded model $M_{\mathcal{S}}$ of $(\mathcal{P} \cup \mathcal{P}_{\mathcal{S}} \cup \mathcal{IC})$ where $\mathcal{P}_{\mathcal{S}}$ is the smallest set of rules containing the fact $A$ if $A \in \mathcal{S}$ and, for all $A' \in \mathcal{AB}_g \backslash \mathcal{S}$ and $A'$ is not the complement of $A$, we have that $A' \leftarrow \textsf{U}$. A set $\mathcal{S}$ is an abductive solution for query $Q$ if $M_{\mathcal{S}} \models Q$.
\end{definition}

\begin{definition}[Transformation for ICs]
For each $\textsf{U} \leftarrow \mathcal{B}_i \in \mathcal{IC}$, we define $\mathcal{A}_i \subseteq \mathcal{B}_i$ as the set of abducibles in the body $\mathcal{B}_i$ of the $i$-th IC, $1 \le i \le m$. We define $\mathcal{IC}'$ as the set of ICs where the body is replaced by $\mathcal{B}_i' = \mathcal{B}_i - \mathcal{A}_i$:
\begin{align*}
    \textsf{U} \leftarrow L_{11}, \dots, L_{1n_1}. 
    \hspace{0.5cm} \cdots \hspace{0.5cm}
    \textsf{U} \leftarrow L_{m1}, \dots, L_{mn_m}.
\end{align*}
The transformation of ICs in $\mathcal{IC}'$ results in the smallest set containing the rules:
\[\textsf{U}^* \leftarrow \textsf{U}^{*1}, \dots, \textsf{U}^{*m}.\]
and
\begin{align*}
    \textsf{U}^{*i} \leftarrow \alpha(L_{i1}), \dots, \alpha(L_{in_i}), assert\_IC(E_{in_i}).
\end{align*}
where $assert\_IC/1$ asserts the facts $ic/1$ to store the list of abducibles obtaining from the $i$-th IC's body and:
\begin{align*}
    \alpha(L_{ij}) = 
		\begin{cases}
	    	l_{ij}(\bar{t}_{ij}, E_{i(j-1)}, E_{ij}) & \text{if $L_{ij} = l_{ij}(\bar{t}_{ij})$} \\
			not\_l_{ij}(\bar{t}_{ij}, E_{i(j-1)}, E_{ij}) & \text{if $L_{ij} = \textit{not }l_{ij}(\bar{t}_{ij})$}
		\end{cases}
\end{align*}
$1 \le j \le n_i$, with $E_{i0} = \mathcal{A}_i$. 
\end{definition}

The predicate $assert\_IC/1$ is a \tabdual system predicate that is only utilized to store the facts of $ic/1$s. This predicate has no direct dependency with any of the tabled predicates. Thus, the use incremental tabling is not necessary. We call $\textsf{U}^{*}$ only once before any of the abduction processes, and in particular, the predicate $assert\_IC/1$.


\begin{example}
    Using the same ICs in Example \ref{ex1}, the result of the new transformation for ICs is as follows:
    \begin{enumerate}
        \item $\textsf{U}^* \leftarrow \textsf{U}^{*1}, \textsf{U}^{*2}.$
        \item $\textsf{U}^{*1} \leftarrow assert\_IC([q(X),r(X)]).$
        \item $\textsf{U}^{*2} \leftarrow u(X,[\;],E), assert\_IC(E).$
    \end{enumerate}
\end{example}

We also modify the query transformation and add a mechanism for ICs testing on each subgoal to filter its solutions early, especially when we have multiple goals in a query.

\begin{definition}[Query Transformation] \label{newic}
Given a query $Q$ as $?- G_1, \dots, G_m.$, we transform it into $\Delta(Q)$, namely $?- \delta(G_1), \dots, \delta(G_m)$ where $\delta$ is defined as follows:
\begin{align*}
    \delta(G_i) = 
		\begin{cases}
	    	g_i(\bar{t}_i, T_{i-1}, T_{i}), test\_IC(T_{i}) & \text{if $G_i = g_{i}(\bar{t}_{i})$} \\
			not\_g_i(\bar{t}_i, T_{i-1}, T_{i}), test\_IC(T_{i}) & \text{if $G_i = not\_g_{i}(\bar{t}_{i})$}
		\end{cases}
\end{align*}
where $T_0 = [\;]$ or other given initial context. We define $test\_IC/1$ as follows
\begin{align*}
    test\_IC(I) \leftarrow forall(ic(X), check\_subset(X,I)).
\end{align*}
where $check\_subset(X,I)$ evaluates to $true$ if $X$ is \textit{not} a subset of $I$ and $forall/2$ is a standard Prolog predicate.
\end{definition}

Essentially, we test the solution in $I$ against each list of abducibles asserted by the $ic/1$ facts. If any of those lists is a subset of a solution, then the solution violates the IC associated with that list of abducibles. Otherwise, the solution is accepted. 

\begin{example} \label{exp:newtr}
    The query $Q$ in Example \ref{ex:query} transforms into  $?-p(0, [\;], O), test\_IC(O)$ and returns only $\mathcal{S} = O = [q(0), q(1), \textit{not } t(0)]$ as a solution. Note that $\{q(0), q(1), \textit{not } t(0), s(0), p(0), \textit{not } u(0)\} \subset M_{\mathcal{S}}$ and thus $M_{\mathcal{S}} \models Q$.
    The query $p(0)$ with input context $r(1)$ correctly returns $false$ since the query is not satisfiable.  Lastly, if we modify the rule $p/1$ into $p(X) \leftarrow q(0), \textit{not } q(1), s(X)$,  the query $Q$ correctly returns $[q(0), \textit{not } q(1), \textit{not } t(0)]$ as a solution instead of failing.
\end{example}


Theorem 3.1 states that with the new transformation for ICs, the size of the whole program resulting from the \tabdual transformation is linear in the size of the input program. 
Suppose $rules(\mathcal{P})$ denotes the number of rules in program $\mathcal{P}$, $|\mathcal{B}_{r_i}|$ denotes the number of literals in the body of rule $r_i$ in $\mathcal{P}$, then, the size of $\mathcal{P}$ is defined as $size(\mathcal{P})$ $size(\mathcal{P}) = \Sigma_{i=1}^{rules(\mathcal{P})} (1 + |\mathcal{B}_{r_i}|)$.

\begin{theorem} \label{th:1}
    Given an abductive logic program $\mathcal{P}$, the set of abducibles $\mathcal{AB}$, and $\tau(\mathcal{P})$ as the resulting program after applying the \tabdual transformation. Then $size(\tau(\mathcal{P})) \le 8 \cdot size(\mathcal{P}) + 4 \cdot |\mathcal{AB}| + 3$.
\end{theorem}


We conduct experiments  using SWI-Prolog on an artificial problem. The experiments aim to compare the table memory usage (in bytes) and the number of logical inferences between programs resulting from different transformation for ICs, viz., the one with the dual program transformation and with the newly proposed transformation with subset checking. The number of logical inferences defined the classical way, i.e., the number of calls performed or rule heads traversed by Prolog,

\begin{experiment} \label{exp:3}
    Given $\mathcal{AB} = \{a, b, c\}$ and $\mathcal{IC} = \{\bot \leftarrow p(X), \textit{not } q(X).\}$. We generate $n$ abductive frameworks $\langle \mathcal{P}_n, \mathcal{AB}, \mathcal{IC} \rangle$ where $\mathcal{P}_n$ is a logic program containing rules of the forms:
    \begin{itemize}
        \item Rules: $p(X) \leftarrow a(X)$; \hspace{0.5cm} $q(X) \leftarrow b(X)$; \hspace{0.5cm} $t(X) \leftarrow c(X).$
        \item Rule: $r \leftarrow a(1), \dots, a(n), \textit{not } q(1), \dots, \textit{not } q(n).$
        \item Rule: $r \leftarrow p(1), \dots, p(n).$
    \end{itemize}
\end{experiment}

\begin{figure}
    \begin{subfigure}{.48\textwidth}
        \centering
        \begin{tikzpicture}
        \begin{axis}[
            xmin=1, xmax=10,
            ymin=100, ymax=5000,
            xtick={1,2,3,4,5,6,7,8,9,10},
            ytick={0,1000,2000,3000,4000,5000},
            legend pos=north west,
            ymajorgrids=true,
            grid style=dashed,
        ]
        \addplot[
            color=blue,
            mark=square,
            ]
            coordinates { (1,482)(2,667)(3,888)(4,1145)(5,1438)(6,1767)(7,2146)(8,2533)(9,2970)(10,3443)
            };
        \addplot[
            color=red,
            mark=triangle,
            ]
            coordinates { (1,227)(2,417)(3,656)(4,953)(5,1317)(6,1757)(7,2282)(8,2901)(9,3623)(10,4457)
            };
            \legend{IC\_Dual,IC\_SubCheck}
        \end{axis}
        \end{tikzpicture}
        \caption{Number of inferences comparison.}
        \label{fig:ex3-1}
    \end{subfigure}
    \begin{subfigure}{.48\textwidth}
        \centering
        \begin{tikzpicture}
        \begin{axis}[
            xmin=1, xmax=10,
            ymin=100, ymax=25000,
            xtick={1,2,3,4,5,6,7,8,9,10},
            ytick={0,5000,10000,15000,20000,25000},
            legend pos=north west,
            ymajorgrids=true,
            grid style=dashed,
        ]
        \addplot[
            color=blue,
            mark=square,
            ]
            coordinates { (1,4536)(2,6592)(3,8648)(4,10704)(5,12760)(6,14816)(7,16872)(8,18928)(9,20984)(10,23040)
            };
        \addplot[
            color=red,
            mark=triangle,
            ]
            coordinates { (1,3760)(2,5816)(3,7872)(4,9928)(5,11984)(6,14040)(7,16096)(8,18152)(9,20208)(10,22264)
            };
            \legend{IC\_Dual,IC\_SubCheck}
        \end{axis}
        \end{tikzpicture}
        \caption{Table memory usage comparison.}
        \label{fig:ex3-2}
    \end{subfigure}
    \caption{Comparison of dual transformation and new transformation for ICs in Experiment \ref{exp:3}.}
    \label{fig:ex3i}
\end{figure}
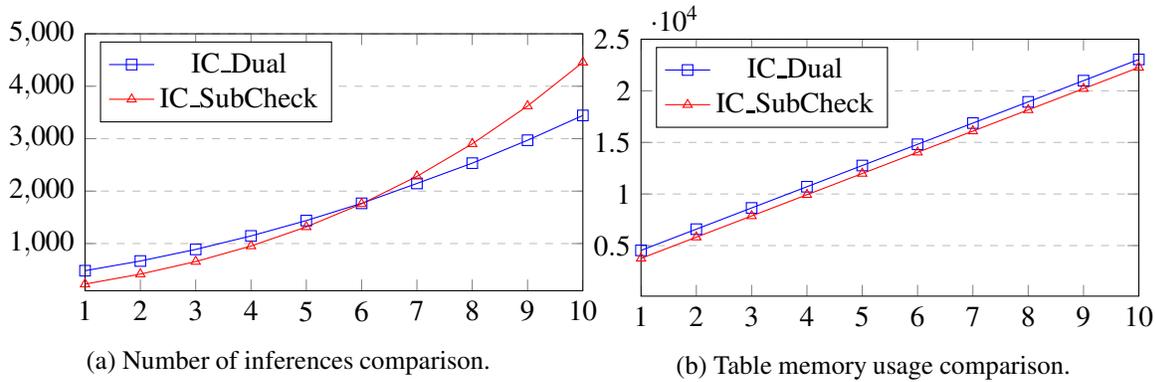

We run the experiments w.r.t. various sizes of abductive solutions as a result of calling the query $?-r, t(1), \dots, t(n)$, $1 \le n \le 10$. 
The rules of $r/0$ are introduced to provide rejected and accepted solutions against the IC. The purpose, as indicated by the query, is to examine the effect of removing solutions that violate the IC early before calling a sequence of $n$ goals $t/1$. The size of the solutions, which varies over $n$, is used as a parameter for evaluating the effect of subset checking over different sizes of abductive solutions. 
The use of $t/1$ in the query is particularly to examine the effect of the new transformation w.r.t. the size of the query. Note that as $n$ increases, the size of the query also increases, as well as the number of subset checking to perform. We define $p/1$ and $q/1$ to slightly add the complexity of the program, where they serve as intermediate predicates before abducing $a$ and $b$. The IC has both positive and negative literals in the body to examine the effect of negative literals occurrences. Note that,  in the dual transformation, the negative literals in the IC's body (here, $not\ q(X)$) are transformed to positive literals, whose predicates are tabled. On the other hand, using the new transformation for ICs, they remain negative and no tabling is involved. 

The result presented in Figure \ref{fig:ex3-1} shows that the number of inferences of the new transformation for ICs using subset checking (\textit{IC\_SubCheck}) starts below that of the dual transformation (\textit{IC\_Dual}), but overtake the \textit{IC\_Dual} mode for $n \ge 6$. It shows that performing subset checking on each subgoal increases the number of inferences during abduction as expected. Even when we prune the solutions early, the number of inferences remains high. However, note that in this problem, the pruning is performed only once during the checking of the first subgoal, and thus does not largely affect the number of inference during abduction. Moreover, the dual transformation for ICs may incorrectly filters the abductive solutions, so the gap between these approaches may still be justified by this remark. Meanwhile, the result in Figure \ref{fig:ex3-2} depicts that the use of the dual transformation for ICs utilizes slightly more table memory, mainly because the predicate $q/1$, obtained from the dual of $not\ q(X)$ in the IC's body is tabled. It shows that the content of the IC's body may affect the table memory usage during abduction.


\section{Optimizing Table Memory Usage}
\label{sect:scale}

In this section, we address the scalability issue during the abduction stage, particularly when dealing with lots and large abductive solutions. We start by introducing a mechanism to reduce the number of tabled solutions by selectively choosing predicates to table. We consider subsequently the simplification of the problem representation to further reduce the table memory usage.

\subsection{Reduced Tabled Predicates}
Considering \textit{every} rule in introducing predicates to table, as described in the transformation for tabling abductive solution (cf. Sect. 2), may certainly result in greedy space consumption, which in turn may hamper the abduction stage to be completed. We propose here a modification of such naive transformation by selecting tabled predicates based on the structure of the rule. That is, we do not table the predicates whose rules contain \textit{only} abducibles or literals that are defined only by facts. Indeed, abducibles in the body can be easily added into the context. On the other hand, facts can be executed straightforwardly since they do not have body. Note that the other predicates whose rules do not meet the above condition still need to be tabled. 

In realizing this modification, we add a procedure to check whether a predicate needs to be tabled based on the above condition. While this reduction is expected to effectively reduce table memory usage, the time and inference needed to process a query may increase depending on the size of the body of the non-tabled predicates' rules, viz., the number of literals in the body. Consider the program in Example \ref{ex1}. Using this new mechanism, the predicate $s/1$ is not tabled since it only has one rule whose body contains only the predicate $t/1$, which is an abducible. However, we still need to table the predicates $p/1$ since, even though its rule's body contains the abducible $q/1$, it also contains $s/1$ which is neither an abducible nor defined only by facts. The predicate $u/1$ is also tabled for similar reason. Note that, by avoiding tabling predicate $s/1$, the same inference is repeated every time $s/1$ is called.

\subsection{Experiments on Artificial Problem}

In this section we conduct experiments using SWI-Prolog to see the growth of the table memory usage (in bytes) and the number of inference calls on processing a query w.r.t. the size of the rule's body of the non-tabled predicates. We compare a program with reduced tabled predicates mechanism ($Reduce$) and the one without such reduction ($Normal$).

\begin{experiment} \label{exp:1}
    Given $\mathcal{AB}_n$ as a set of $n$ abducibles $\{a_1,\dots,a_n\}$. We generate an abductive framework $\langle \mathcal{P}_n, \mathcal{AB}_n, \emptyset \rangle$ where $\mathcal{P}_n$ is a logic program containing rules of the forms:
    \begin{itemize}
        \item Facts $b_i(X).$ for $1 \le i \le n$.
        \item Rules $p_i(X) \leftarrow a_1(X),\dots,a_i(X), b_1(X),\dots,b_i(X).$ for $1 \le i \le n$.
        \item Rules $q_i(X) \leftarrow p_1(X), \dots, p_i(X).$ for $1 \le i \le n$.
    \end{itemize}
    For this experiment, we use $n = 10$.
\end{experiment}

\begin{figure}
    \begin{subfigure}{.48\textwidth}
        \centering
        \begin{tikzpicture}
        \begin{axis}[
            xmin=1, xmax=10,
            ymin=100, ymax=1200,
            xtick={1,2,3,4,5,6,7,8,9,10},
            ytick={0,200,400,600,800,1000,1200},
            legend pos=north west,
            ymajorgrids=true,
            grid style=dashed,
        ]
        \addplot[
            color=blue,
            mark=square,
            ]
            coordinates { (1,180)(2,300)(3,366)(4,443)(5,532)(6,634)(7,759)(8,881)(9,1028)(10,1192)
            };
        \addplot[
            color=red,
            mark=triangle,
            ]
            coordinates { (1,149)(2,238)(3,273)(4,319)(5,377)(6,448)(7,533)(8,633)(9,749)(10,882)
            };
            \legend{Normal,Reduce}
        \end{axis}
        \end{tikzpicture}
        \caption{Number of inferences on non-incremental call.}
        \label{fig:ex1-1}
    \end{subfigure}
    \begin{subfigure}{.48\textwidth}
        \centering
        \begin{tikzpicture}
        \begin{axis}[
            xmin=1, xmax=10,
            ymin=100, ymax=45000,
            xtick={1,2,3,4,5,6,7,8,9,10},
            ytick={0,5000,15000,25000,35000,45000},
            legend pos=north west,
            ymajorgrids=true,
            grid style=dashed,
        ]
        \addplot[
            color=blue,
            mark=square,
            ]
            coordinates { (1,1872)(2,3224)(3,4864)(4,6792)(5,9008)(6,11512)(7,14304)(8,17384)(9,20752)(10,24408)
            };
        \addplot[
            color=red,
            mark=triangle,
            ]
            coordinates { (1,1096)(2,1384)(3,1672)(4,1960)(5,2248)(6,2536)(7,2824)(8,3112)(9,3400)(10,3688)
            };
            \legend{Normal,Reduce}
        \end{axis}
        \end{tikzpicture}
        \caption{Table memory usage on non-incremental call.}
        \label{fig:ex1-2}
    \end{subfigure}
    \begin{subfigure}{.48\textwidth}
        \centering
        \begin{tikzpicture}
        \begin{axis}[
            xmin=1, xmax=10,
            ymin=100, ymax=1200,
            xtick={1,2,3,4,5,6,7,8,9,10},
            ytick={0,200,400,600,800,1000,1200},
            legend pos=north west,
            ymajorgrids=true,
            grid style=dashed,
        ]
        \addplot[
            color=blue,
            mark=square,
            ]
            coordinates { (1,180)(2,191)(3,161)(4,204)(5,257)(6,321)(7,397)(8,486)(9,589)(10,707)
            };
        \addplot[
            color=red,
            mark=triangle,
            ]
            coordinates { (1,149)(2,159)(3,130)(4,176)(5,234)(6,305)(7,390)(8,490)(9,606)(10,739)
            };
            \legend{Normal,Reduce}
        \end{axis}
        \end{tikzpicture}
        \caption{Number of inferences on incremental call.}
        \label{fig:ex1-3}
    \end{subfigure}
    \begin{subfigure}{.48\textwidth}
        \centering
        \begin{tikzpicture}
        \begin{axis}[
            xmin=1, xmax=10,
            ymin=100, ymax=45000,
            xtick={1,2,3,4,5,6,7,8,9,10},
            ytick={0,5000,15000,25000,35000,45000},
            legend pos=north west,
            ymajorgrids=true,
            grid style=dashed,
        ]
        \addplot[
            color=blue,
            mark=square,
            ]
            coordinates { (1,1872)(2,4000)(3,6704)(4,9984)(5,13840)(6,18272)(7,23280)(8,28864)(9,35024)(10,41760)
            };
        \addplot[
            color=red,
            mark=triangle,
            ]
            coordinates { (1,1096)(2,2160)(3,3512)(4,5152)(5,7080)(6,9296)(7,11800)(8,14592)(9,17672)(10,21040)
            };
            \legend{Normal,Reduce}
        \end{axis}
        \end{tikzpicture}
        \caption{Table memory usage on incremental call.}
        \label{fig:ex1-4}
    \end{subfigure}
    \caption{Number of inferences and table memory usage of Experiment \ref{exp:1}.}
    \label{fig:ex1}
\end{figure}
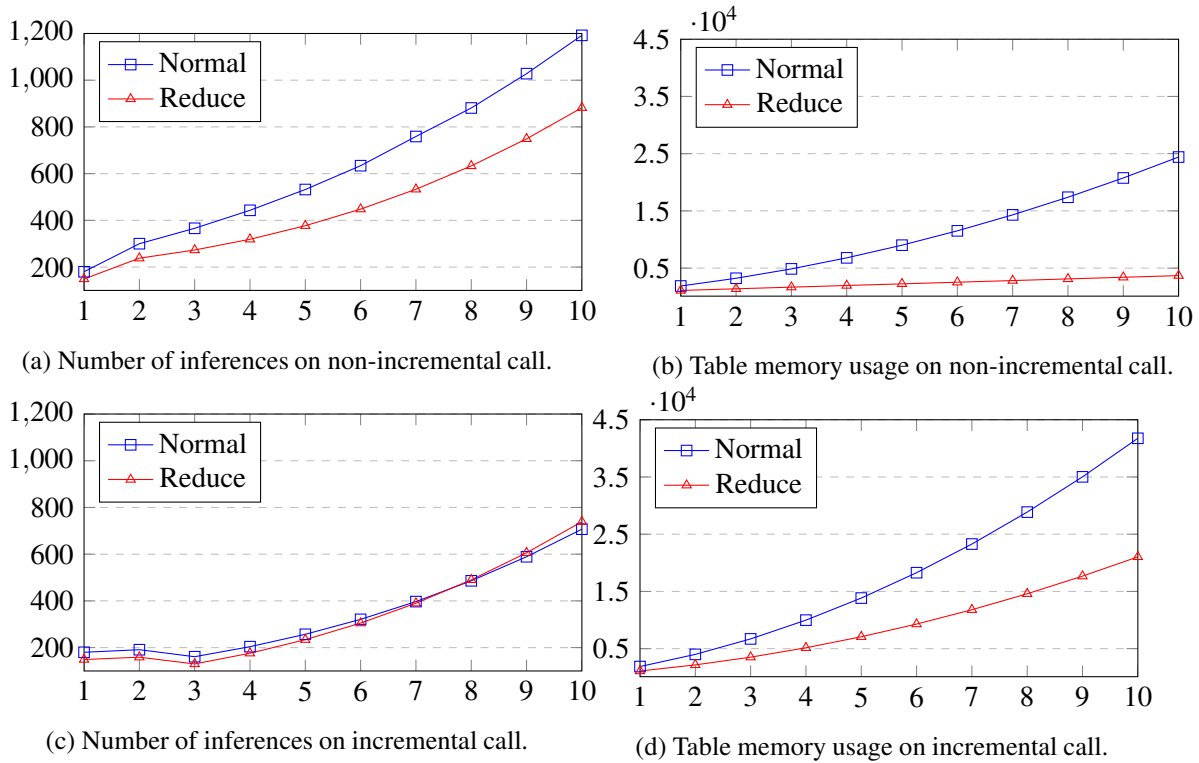

We run these experiments for various size of the predicate $p_i$'s body for $1 \le i \le n$ using the query $?- q_i(1)$ both in $Normal$ and $Reduce$ mode. The rules of $p_i/1$ are defined only by abducibles $a_i$ and predicate $b_i/1$; the latter is defined only by facts. We define $q_i/1$ for the query that can reuse the solutions of $p_1/1$ to $p_i/1$ if they are tabled. The size of $q_i/1$'s body that varies w.r.t. $i$ becomes a parameter to analyse the number of inferences and the table memory usage. For example, the query $q_2(1)$ will call the subgoal $p_1(1)$ and $p_2(1)$, while query $q_3(1)$ will additionally call $p_3(1)$. In $Normal$ mode, i.e., without reducing the tabled predicates, each time the rule $p_i(X)$ is called, we table its solutions which can be reused in subsequent calls. However, in the $Reduce$ mode, those results are not tabled and have to be recomputed.

We run these experiments on two different query calling cases. We define the \textit{incremental call} of the queries, in which we make use of the previously tabled predicate $p_i(X)$ in normal mode. For instance, the call of query $q_2(1)$ makes use of the table produced during the call of query $q_1(1)$, which has previously been executed. On the other hand, in the  \textit{non-incremental call} of each query, the resulting table is abolished after each query call. As an example, after the call of $q_1(1)$, we abolish the table before executing the next query, $q_2(1)$.

On Figure \ref{fig:ex1-1}, there is a decrease of inference calls on non-incremental call in \textit{Reduce} mode compared to that of \textit{Normal} mode. It shows that by not tabling some predicates, we can reduce the number of inferences, especially those on the tabling operations. In Figure \ref{fig:ex1-2}, the table memory usage of \textit{Reduce} mode is significantly lower compared to that of \textit{Normal} mode. The gap is larger as $i$ increases, since, for example, query $q_{10}(1)$ creates the tables for $p_1(1)$ until $p_{10}(1)$ in addition to $q_{10}/1$. That is not the case on \textit{Reduce} mode, since $p_i/1$ is not tabled, and the program only produces table for $q_{10}/1$. The graph thus shows only little increases over $i$. 

On the incremental call, Figure \ref{fig:ex1-3} shows that the number of inferences for \textit{Reduce} mode starts below the \textit{Normal} mode but, at some point, overtakes that of \textit{Normal} mode. It happens since, in the \textit{Normal} mode, the previously tabled predicates are used on the next query call consecutively. For example, when calling the query $q_3(1)$, the program can reuse the table of $p_1(1)$ and $p_2(1)$ that is produced during the execution of the previous query $q_1(1)$ and $q_2(1)$. This is not the case for the \textit{Reduce} mode since on every call, it must execute $p_i/1$ again as they are not tabled. Nevertheless, the table memory usage in Figure \ref{fig:ex1-4} for \textit{Reduce} mode is still lower than that of \textit{Normal} mode. The increase in table memory usage between Figure \ref{fig:ex1-2} and \ref{fig:ex1-4} happens since the table for all $q_i/1$ from the previous query execution is not deleted before the execution of the next query during the incremental call. While in non-incremental call, those tables are abolished between each call.


\subsection{Experiments on Real World Problem}

These experiments are performed on an abduction problem in chemoprevention \cite{LK12}. Abduction is enacted to study genes that affect cancer cells activation or inactivation, with the knowledge on active or inactive cells and their propagation pathways. The problem itself is challenging since the program contains a large number of facts and rules with intricate relationships of genes and cancer cells.

\subsubsection{Experimental Results}

The abduction in these experiments are conducted with answer subsumption to compute only minimal explanations. Furthermore, we shall compare the scalability of the proposed approach in this paper and the previous work in \cite{PS17} with answer subsumption only. In this evaluation, we use XSB Prolog since there is a difference on partial order mechanism between answer subsumption in XSB Prolog and SWI Prolog implementation. XSB Prolog can provide multiple minimal abductive solutions in case they are incomparable w.r.t. the subset operation, while SWI Prolog can only provide a single solution.

\begin{experiment} \label{exp:2}
    The main query comprises eight subgoals:
    \begin{align*}
	    & active(phase0, aif),\; active(phase0, endo\_g),\; inactive(phase0, caspase9), \\ 
	    & inactive(phase0, caspase6),\; inactive(phase0, bcl2),\; inactive(phase0, caspase7), \\
	    & inactive(phase0, akt),\; inactive(phase0, xiap).
    \end{align*}
    The ICs of this problem is as follows:
    \begin{enumerate}
        \item $\leftarrow drug\_induced(phase0, drug, Gene), drug\_inhibited(phase0, drug, Gene).$
        \item $\leftarrow drug\_induced(phase0, drug, apoptosis).$
    \end{enumerate}
    where $drug\_induced/3$ and $drug\_inhibited/3$ are both abducibles.
\end{experiment}

The objective is to analyse the effects on the table memory usage (in bytes) when we consider answer subsumption only (\textit{AS}), a combination with reduced tabled predicates (\textit{AS+Reduce}), as well as a further combination with the new transformation for ICs by subset checking  (\textit{AS + Reduce + SC}). We run the query, starting with only the first subgoal, and continuing to harder queries with more subgoals, until all eight subgoals are executed. However, unlike the experiment in \cite{PS17}, we do so non-incrementally, i.e., we abolish the table between each query execution. We also set the timeout for 25200 seconds (7 hours).

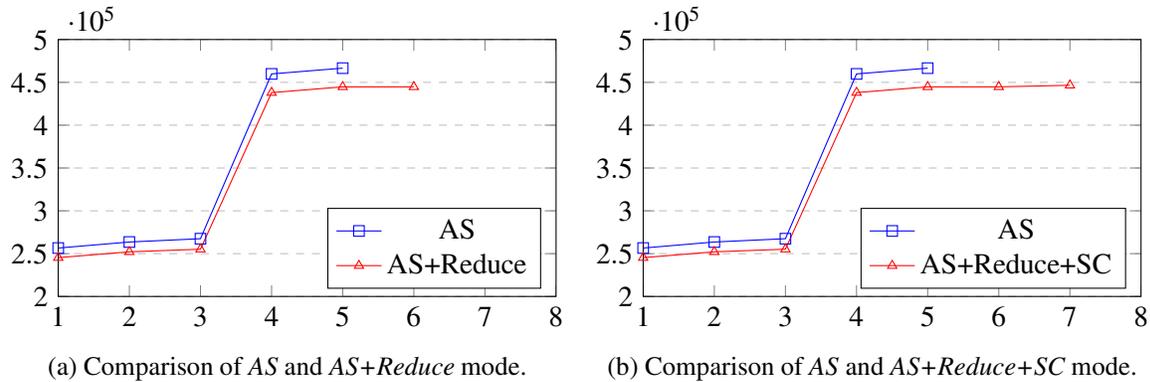
\begin{figure}
    \begin{subfigure}{.48\textwidth}
        \centering
        \begin{tikzpicture}
        \begin{axis}[
            xmin=1, xmax=8,
            ymin=200000, ymax=500000,
            xtick={1,2,3,4,5,6,7,8},
            ytick={200000,250000,300000,350000,400000,450000,500000},
            legend pos=south east,
            ymajorgrids=true,
            grid style=dashed,
        ]
        \addplot[
            color=blue,
            mark=square,
            ]
            coordinates { (1,256560)(2,263588)(3,267452)(4,459860)(5,466600)
            };
        \addplot[
            color=red,
            mark=triangle,
            ]
            coordinates { (1,245332)(2,252116)(3,255192)(4,438016)(5,444748)(6,444748)
            };
            \legend{AS,AS+Reduce}
        \end{axis}
        \end{tikzpicture}
        \caption{Comparison of \textit{AS} and \textit{AS+Reduce} mode.}
        \label{fig:ex2-1}
    \end{subfigure}
    \begin{subfigure}{.48\textwidth}
        \centering
        \begin{tikzpicture}
        \begin{axis}[
            xmin=1, xmax=8,
            ymin=200000, ymax=500000,
            xtick={1,2,3,4,5,6,7,8},
            ytick={200000,250000,300000,350000,400000,450000,500000},
            legend pos=south east,
            ymajorgrids=true,
            grid style=dashed,
        ]
        \addplot[
            color=blue,
            mark=square,
            ]
            coordinates { (1,256560)(2,263588)(3,267452)(4,459860)(5,466600)
            };
        \addplot[
            color=red,
            mark=triangle,
            ]
            coordinates { (1,245332)(2,252116)(3,255192)(4,438048)(5,444748)(6,444748)(7,446584)
            };
            \legend{AS,AS+Reduce+SC}
        \end{axis}
        \end{tikzpicture}
        \caption{Comparison of \textit{AS} and \textit{AS+Reduce+SC} mode.}
        \label{fig:ex2-3}
    \end{subfigure}
    \caption{Comparison of table memory usage in Experiment \ref{exp:2}}
    \label{fig:ex2t}
\end{figure}

The comparison of the resulting table memory usage is presented in Figure \ref{fig:ex2t}. We see that, in Figure 3a, while \cite{PS17} manages to finish all queries incrementally, they are still unable to finish the execution for more than five subgoals non-incrementally. By reducing the tabled predicates, we manage to execute six subgoals. Unfortunately, for this particular problem, the decrease of the table memory usage still cannot help in finishing all subgoals. In Figure \ref{fig:ex2-3}, using \textit{AS+Reduce+SC} mode, we manage to complete seven subgoals. The reason is that the program in \textit{AS} mode tries to add several sets of abducibles into the solutions while the program on \textit{AS+Reduce+SC} does not, since it uses the new transformation for ICs. Thus, it produces smaller and less number of solutions compared to the program in \textit{AS} mode, that produces at most twice as much solutions since the first IC in the problem contains two literals. We can also see that the table memory usage with the proposed approach is less than the \textit{AS} mode. Even so, the difference is relatively small, which may indicate that non-tabled predicates are not called as often as the tabled ones. 

\subsubsection{Simplification of Domain-Specific Model Representation}

Since the executions of all subgoals is still not possible, we explore another approach to try and finish all subgoals executions. We further simplify the representation of the problem by eliminating the first argument of the goals, viz., \textit{phase0}, as there is no relationship between \textit{phase0} and others in the given knowledge base. This argument can therefore safely be eliminated from the model representation. In the case where many phases of experiments occur, we can simply run the abductive program several times while adjusting the experiment-specific part of the program to each independent phase. The simplification is performed to decrease the table memory usage further.

\begin{experiment} \label{exp:5}
   Since we remove phase from the program, the main query is redefined as:
   \begin{align*}
	    & active(aif),active(endo\_g),inactive(caspase9),inactive(caspase6),\\
	    & inactive(bcl2),inactive(caspase7),inactive(akt),inactive(xiap).
    \end{align*}
    The ICs are redefined accordingly.
\end{experiment}

\begin{figure}
    \begin{subfigure}{.48\textwidth}
        \centering
        \begin{tikzpicture}
        \begin{axis}[
            xmin=1, xmax=8,
            ymin=200000, ymax=500000,
            xtick={1,2,3,4,5,6,7,8},
            ytick={200000,250000,300000,350000,400000,450000,500000},
            legend pos=south east,
            ymajorgrids=true,
            grid style=dashed,
        ]
        \addplot[
            color=blue,
            mark=square,
            ]
            coordinates { (1,256560)(2,263588)(3,267452)(4,459860)(5,466600)
            };
        \addplot[
            color=red,
            mark=triangle,
            ]
            coordinates { (1,245332)(2,252116)(3,255192)(4,438016)(5,444748)(6,444748)
            };
        \addplot[
            color=black,
            mark=diamond,
            ]
            coordinates { (1,233512)(2,239188)(3,242004)(4,399420)(5,405100)(6,405132)
            };
            \legend{\footnotesize AS,\footnotesize AS+Reduce,\footnotesize AS+Simp+Reduce}
        \end{axis}
        \end{tikzpicture}
        \caption{Table memory usage comparison with dual transformation for ICs.}
        \label{fig:ex4-1}
    \end{subfigure}
    \begin{subfigure}{.48\textwidth}
        \centering
        \begin{tikzpicture}
        \begin{axis}[
            xmin=1, xmax=8,
            ymin=200000, ymax=500000,
            xtick={1,2,3,4,5,6,7,8},
            ytick={200000,250000,300000,350000,400000,450000,500000},
            legend pos=south east,
            ymajorgrids=true,
            grid style=dashed,
            width=7.6cm,
        ]
        \addplot[
            color=blue,
            mark=square,
            ]
            coordinates { (1,256560)(2,263588)(3,267452)(4,459860)(5,466600)
            };
        \addplot[
            color=red,
            mark=triangle,
            ]
            coordinates { (1,245332)(2,252116)(3,255192)(4,438048)(5,444748)(6,444748)(7,446584)
            };
        \addplot[
            color=black,
            mark=diamond,
            ]
            coordinates { (1,233512)(2,239188)(3,242004)(4,399388)(5,405100)(6,405100)(7,406696)
            };
            \legend{\footnotesize AS,\scriptsize AS+Reduce+SC,\scriptsize AS+Simp+Reduce+SC}
        \end{axis}
        \end{tikzpicture}
        \caption{Table memory usage comparison with new transformation for ICs.}
        \label{fig:ex4-2}
    \end{subfigure}
    \caption{Comparison of table memory usage between initial and simplified model representation.}
    \label{fig:ex4t}
\end{figure}
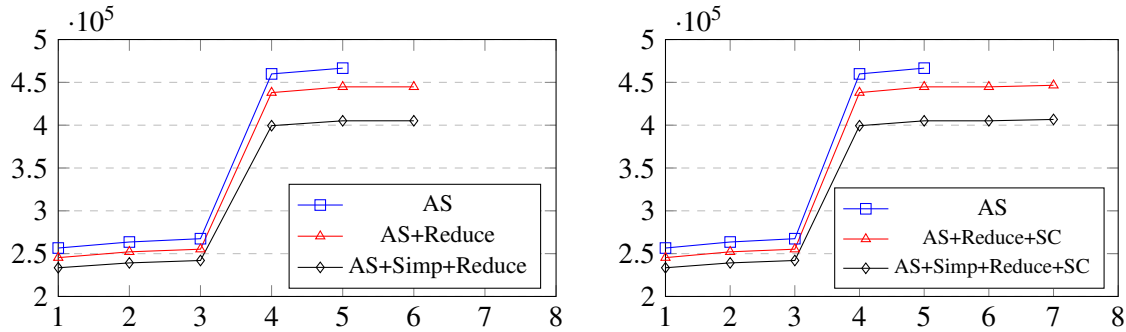

The objective and scenario of these experiments are similar with Experiment \ref{exp:2}. However, the query is adjusted as in Experiment \ref{exp:5}. We also consider the effect on the representation simplification (\textit{Simp}). Based on Figure \ref{fig:ex4t}, we can see that the the simplification of the model can successfully reduce the table memory usage further, since the arity of each predicate has been reduced by one, and thus reducing the amount of memory needed to table them. However, even with the decrease of table memory usage, the execution of eight subgoals is still not possible even with the new transformation for ICs on \textit{AS+Simp+Reduce+SC} mode. It shows that the number of abductive solutions generated for eight subgoals may still be too numerous for the program to terminate within the allocated time of 25200 seconds.

\section{Concluding Remarks}
\label{sect:cfw}

In this paper, we propose a new transformation for ICs to warrant their correct application in filtering solutions by utilizing subset checking as a way to ensure that the abductive solutions adheres to the ICs, instead of using the dual program transformation. However, this approach may potentially increase the number of inferences during abduction, as presented in the experiments. We also introduce a mechanism to reduce the number of tabled solutions and simplify the problem representation to further reduce the table memory usage. Our experiments show that by reducing the number of tabled predicates, we can decrease the table memory usage during the abduction stage. As a trade off, the number of inferences may increase if the body of the non-tabled predicate is large. Furthermore, by simplifying the representation of the problem, the table memory usage can be decreased further. However, this approach is problem specific, and requires an understanding of the problem domain in order to be implemented successfully. In our experiment on Abduction in Chemo-prevention, we also managed to finish the abduction process non-incrementally up to seven subgoals using the reduction and the new transformation for ICs. The previous results in \cite{PS17} can only finish five subgoals non-incrementally. The results for eight subgoals are still unavailable due to the size of the abductive solutions generated during the abduction process.

The idea of tabling abductive solutions from one abductive context to be reused in other contexts (contextual abduction) is presented by \textsc{Abdual} \cite{AP04}, as a theory for abduction over Well-Founded semantics. In \cite{SP15}, the technique for the program transformation from abductive normal logic programs into tabled logic programs is presented employing the dual transformation of \textsc{Abdual}. 
Several works to improve \tabdual have been proposed in \cite{IS19,PS17} to reduce the tabling memory usage during the abduction process. However, both implementation  utilize some tabling features and do not make any adjustment in the program transformation. Moreover, there are several issues regarding those implementations as we previously stated.

We use the same benchmark as in \cite{IS19,PS17} for evaluating our proposed approaches, which use the chemoprevention abductive program previously described in \cite{LK12}. This abduction system is build on top of the A-system \cite{NK01}. Given a query $Q$, the A-system computes an extension of the abducibles such that it entails $Q$ and the ICs. We cannot compare our result directly to the result in \cite{LK12}. One problem that prevents the execution of all subgoals is the large number and size of the abductive solutions. In \cite{LK12}, the size of the solutions are reduced by limiting the search depth of the A-system, showing that even in low depth searches, the quality of the provided results are satisfactory. Currently, we cannot replicate it in both SWI-Prolog and XSB-Prolog, since both have no functionality to limit search depth. Thus, it may be interesting to explore other ways to limit the size of the solutions, such as the bounded rationality approach in XSB-Prolog. 

We may also attempt to selectively decrease the number of tabled solutions further, e.g., by removing redundant solutions. In this paper, we implement subset minimality for the answer subsumption criteria in selecting preferred solutions. Another approach that can be explored is by using minimal cardinality criteria. Furthermore, it may also be interesting to solve different challenging real-life problems using \tabdual as other benchmarks. Lastly, the use of meta-interpreter which can be unfolded w.r.t. the interpreted program to generate an executable program free of the meta-interpreter overhead can also be examined further.

\paragraph{Acknowledgement}
This work was supported by the PUTI grant NKB-847/UN2.RST/HKP.05.00/2020 funded by the Directorate of Research \& Development Universitas Indonesia

\bibliographystyle{eptcs}
\bibliography{generic}
\end{document}